\documentclass[authoryear]{elsarticle}

\usepackage{amsfonts}
\usepackage{amssymb}
\usepackage{graphicx}
\usepackage{amsmath}
\usepackage{tabularx}
\usepackage{algorithm}
\usepackage[noend]{algpseudocode}
\usepackage[utf8]{inputenc}
\usepackage[toc,page]{appendix}
\usepackage{multirow}
\usepackage{tablefootnote}
\newdefinition{definition}{Definition}

\usepackage{color,soul,booktabs}
\usepackage[colorlinks,
            linkcolor=blue,
            citecolor=blue,
            urlcolor=magenta,
            linktocpage,
            plainpages=false]{hyperref}
\setulcolor{blue}

\DeclareMathOperator{\uar}{{\relbar\mkern-9mu\relbar}}
\newcommand{\Prob}{\operatorname{P}}
\newcommand{\Val}{\mathit{Val}}
\newcommand{\Ga}{\mathit{Gamma}}
\newcommand{\Dir}{\mathit{Dir}}
\newcommand{\given}{\operatorname{\mid}}
\newcommand{\N}{\mathcal{N}}
\newcommand{\G}{\mathcal{G}}
\newcommand{\D}{\mathcal{D}}
\newcommand{\U}{\mathbf{U}}

\renewcommand{\u}{\mathbf{u}}
\newcommand{\X}{\mathbf{X}}
\newcommand{\Q}{\mathbf{Q}}
\newcommand{\q}{\mathbf{q}}
\newcommand{\btheta}{\boldsymbol{\theta}}
\newcommand{\ML}{\operatorname{ML}}

\newcommand{\QXkU}{\Q_{X_k \given \U}}
\newcommand{\QXku}{\Q_{X_k \given \u}}
\newcommand{\qu}[1]{q_{#1 \given \u}}
\newcommand{\qXkU}{\q_{X_k \given \U}}
\newcommand{\qXku}{\q_{X_k \given \u}}

\newcommand{\btXku}{\btheta_{X_k \given \u}}
\newcommand{\btXkU}{\btheta_{X_k \given \U}}
\newcommand{\tiu}{\tau_{x_i \given \u}}
\newcommand{\aiu}{\alpha_{x_i \given \u}}
\newcommand{\aiju}{\alpha_{x_i x_j \given \u}}
\newcommand{\Sij}{\mathbf{S}_{X_i X_j}}
\newcommand{\s}{\mathbf{s}}
\newcommand{\Ms}{M_{xx' \given \s}}
\newcommand{\Mys}{M_{xx' \given y,\s}}
\newcommand{\indep}{\perp\hspace{-0.21cm}\perp}

\begin{document}

    \begin{frontmatter}
        \title{A Constraint-Based Algorithm for the Structural Learning of Continuous-Time Bayesian Networks}
        \author[1]{Alessandro Bregoli$^\dag$\footnote{$\dag$ Corresponding author}}\ead{a.bregoli1@campus.unimib.it}
        \author[2]{Marco Scutari}\ead{scutari@idsia.ch}
        \author[1]{Fabio Stella}\ead{fabio.stella@unimib.it}
        
        \address[1]{Department of Informatics, Systems and Communication,\\ University of Milan-Bicocca, Milan, Italy}
        \address[2]{Istituto Dalle Molle di Studi sull'Intelligenza Artificiale (IDSIA), Lugano, Switzerland}
        \begin{abstract}
            Dynamic Bayesian networks have been well explored in the literature as discrete-time models: however, their continuous-time extensions have seen comparatively little attention.
            In this paper, we propose the first constraint-based algorithm for learning the structure of continuous-time Bayesian networks.
            We discuss the different statistical tests and the underlying hypotheses used by our proposal to establish conditional independence. Furthermore, we analyze and discuss the computational complexity of the best and worst cases for the proposed algorithm.
            Finally, we validate its performance using synthetic data, and we discuss its strengths and limitations comparing it with the score-based structure learning algorithm from \citet{nsk03}.
            We find the latter to be more accurate in learning networks with binary variables, while our constraint-based approach is more accurate with variables assuming more than two values. Numerical experiments confirm that score-based and constraint-based algorithms are comparable in terms of computation time. 
        \end{abstract}
        
    \end{frontmatter}

\section{Introduction}
\label{sec:intro}

Multivariate time-series data are becoming increasingly common in many domains such as healthcare, medicine, biology, finance, telecommunications, social networks, \mbox{e-commerce}, and homeland security.
Their size, { (the number of observations)}, and dimensionality, { (the number of variables), are} set to continue to increase in the future, requiring automated algorithms to discover the probabilistic structure {of the underlying data-generating process} and to predict their trajectories over time.

In this paper we focus on the problem of learning the structure of continuous-time Bayesian
networks \citep[CTBNs;][]{NSK02} from data. 
This type of probabilistic graphical model has been successfully used to reconstruct transcriptional regulatory networks from time-course gene expression data \citep{AVPMZS16}, to model the presence of people at their computers \citep{nh03}, and to detect network intrusion \citep{xs08}. Recently CTBNs have been extended to model; non-stationary time series \citep{ViSt}, survival times with arbitrary distributions \citep{pmlr-v119-engelmann20a}, and situations where a system’s state variables could be influenced by occurrences of external events such as interventions \citep{BHal}.
The literature implements CTBN structure learning using score-based algorithms that maximize the Bayesian-Dirichlet equivalent (BDe) metric. 

The main contributions of this paper are:
\begin{itemize}
	\item the design of the first constraint-based algorithm for the structure learning of CTBNs, which we call
	  \textit{Continuous-Time PC} (CTPC);
	\item the definition of suitable test statistics to assess conditional independence in CTBNs;
	\item the time complexity analysis of best and worst cases of the CTPC algorithm;
	\item an empirical performance comparison between score-based algorithm and our proposal.
\end{itemize}
The rest of the paper is organized as follows. 
We introduce CTBNs and the associated score-based structure learning algorithms in Section~\ref{sec:ctbns}.
After a brief introduction to constraint-based learning for BNs (Section~\ref{sec:bnsl}), we propose a constraint-based algorithm and the associated conditional independence tests for CTBNs in Section~\ref{sec:proposal}, followed an analysis of their time complexity analysis in Section~\ref{sec:bigO}.
We then compare score-based and constraint-based approaches in Section~\ref{sec:simulations}, summarizing our conclusions in Section~\ref{sec:conclusions}.

\section{Continuous-Time Bayesian Networks}
\label{sec:ctbns}

CTBNs are a class of probabilistic graphical models that combines Bayesian networks \citep[BNs;][]{koller} and homogeneous Markov processes to model discrete-state continuous-time dynamical systems \citep{NSK02}. Compared to their discrete-time counterpart, dynamic Bayesian networks (DBNs), they can efficiently model domains like those mentioned above in which variables evolve at different time granularities.

{ Several approximate inference algorithms for filtering and smoothing in CTBNs have been proposed in the literature  \citep{BuCe,TalElFri, NoKoSHbis, FanXS10}, while complexity of exact and approximate inference in CTBNs has been shown to be NP-hard \citep{STURLA14}.
.}

{ CTBNs are particularly interesting because they allow us to discuss and reason more naturally about systems in which: {\em i}) events, measurements, or durations are irregularly spaced, {\em ii}) rates vary by orders of magnitude, or {\em iii}) the duration of continuous measurements need to be expressed explicitly.
Other models that can represent discrete-state continuous time processes include Poisson networks \citep{rajaram05a}, cascade of Poisson process model \citep{Simma}, piecewise-constant intensity models \citep{GunaWa}, forest-based point processes \citep{Weiss13}, and graphical models for marked point processes \citep{didelez2008graphical}. 

The main advantages of CTBNs, when compared to the models above, are: 
\begin{itemize}
    \item their graphical structure makes it possible to understand and explain the underlying stochastic process they model,
    \item prior knowledge from domain experts can be integrated in structure learning by blacklisting and whitelisting arcs.
\end{itemize}
However, CTBNs have a number of limitations: they do not allow modeling
continuous state variables; they do not model point events very well, particularly if there are non-temporal values associated with the events; and structure learning is computationally challenging.}

\subsection{Definitions and Notations}
\label{sec:notation}

CTBNs are based on finite-state continuous-time homogeneous Markov processes, that is, stochastic processes  in which  the  transition  intensities do not depend on time.
Let $X$ be a random variable whose state can take $m$ discrete values $Val(X)=\{x_1, ..., x_m \}$.    $X$ changes its state continuously over time $t$. A homogeneous Markov process $X(t)$ is described with its \textit{intensity matrix}:
\begin{align*}
  &\Q_{X} = 
  \begin{bmatrix}
    -q_{x_1}       &  q_{x_1 x_2}   & \cdots &  q_{x_1 x_m} \\
      q_{x_2 x_1}  & -q_{x_2}       & \cdots &  q_{x_2 x_m} \\
      \vdots 		&  \vdots         & \ddots &  \vdots \\
      q_{x_m x_1}  &  q_{x_m x_2}   & \cdots & -q_{x_m}
  \end{bmatrix}.
\end{align*}

The matrix $\Q_{X}$ allows us to describe the transient behaviour of the random variable $X$.   If at time $t=0$ the random variable is in state $x_i$,
then it stays there for an amount of time that is distributed as an exponential random variable with parameter $q_{x_i}$.  Therefore, the probability density function $f\left( t\right)$ and the distribution function $F\left( t\right)$ for the random variable $X(t)$ to remain in state $x_i$ are
\begin{align*}
  &f\left( t\right) = q_{x_i}\exp \left( -q_{x_i} t\right)&
  &\text{and}&
  &F\left( t\right) = 1-\exp \left( -q_{x_i} t\right)
\end{align*}
\noindent where $t\geq 0$. The expected time of transitioning from state $x_i$ is $1/q_{x_i}$; when transitioning from state $x_i$ the random variable $X$ shifts to state $x_j$ with probability $q_{x_{i}x_{j}}/q_{x_i}$.

However, the size of the intensity matrix $\Q_{X}$, which corresponds to the state space of the Markov process, grows exponentially with the number of
variables and with their cardinality. This makes the above representation infeasible for models including more than a very small
number of variables. Therefore, we introduce conditional Markov processes  in order to model larger Markov processes as CTBNs.

A \textit{conditional Markov process} is an inhomogeneous Markov process in which, for any given random variable, the intensities are a function of the current values of a particular set of other variables, which also evolve as Markov processes. Therefore, intensities vary over time but not as a function of time. 
To clarify how a conditional Markov process is described, let $X$ be a random variable whose domain is $\Val(X)=\{x_1, ..., x_m \}$  and assume that it evolves  as a
Markov process $X(t)$. Furthermore, assume that the dynamics of $X(t)$ are conditionally dependent from a set $\U$ of random variables evolving over time.     Then the
dynamics of $X(t)$ can be fully described by means of a conditional intensity matrix (CIM), which can be written as follows:
\begin{align*}
  &\Q_{X \given \U} = 
  \begin{bmatrix}
    -q_{x_1 \given\u}       &  q_{x_1 x_2 \given \u}   & \cdots &  q_{x_1 x_m \given \u} \\
      q_{x_2 x_1 \given \u}  & -q_{x_2 \given \u}       & \cdots &  q_{x_2 x_m \given \u} \\
      \vdots 		&  \vdots         & \ddots &  \vdots \\
      q_{x_m x_1 \given \u}  &  q_{x_m x_2 \given \u}   & \cdots & -q_{x_m \given \u}
  \end{bmatrix}.
\end{align*}

\noindent A CIM is a set of intensity matrices, one intensity matrix for each instance of values $\u$ to the set of variables $\U$.

A CTBN models a stochastic process over a structured state space for a set of random variables $\X = \{X_1, X_2, \ldots, X_n\}$, where each $X_k\in \X$ takes values over a finite domain $\Val(X_k)$.
It encodes such a process in a compact form by factorizing its dynamics into local continuous-time Markov processes that depend on a limited set of states. 
\begin{definition} 
\citep{NSK02}
A CTBN $\N$ over $\X$ is characterized by two components: 
\begin{itemize}
    \item An initial distribution $\Prob^{0}(\X)$, specified as a BN over $\X$.
    \item A continuous-time transition model specified as:
    \begin{itemize}
        \item a directed (possibly cyclic) graph $\G$ whose nodes correspond to the $X_k \in \X$;
        \item a conditional intensity matrix $\QXkU$ for each $X_k$. 
    \end{itemize}
\end{itemize}
\end{definition}
The conditional intensity matrix (CIM) $\QXkU$ consists of the set of intensity matrices 
\begin{align*}
  &\QXku = 
  \begin{bmatrix}
    -\qu{x_1}       &  \qu{x_1 x_2}   & \cdots &  \qu{x_1 x_m} \\
      \qu{x_2 x_1}  & -\qu{x_2}       & \cdots &  \qu{x_2 x_m} \\
      \vdots 		&  \vdots         & \ddots &  \vdots \\
      \qu{x_m x_1}  &  \qu{x_m x_2}   & \cdots & -\qu{x_m}
  \end{bmatrix},&
  &m = |\Val(X_k)|,\footnotemark
\end{align*}
one for each possible configuration $\u$ of the parents $\U$ of $X_k$ in $\G$.

\footnotetext{For simplicity of notation, and without loss of generality, we omit the $k$ subscript from $m$ that implies that each $X_k$ may have a different domain.}
The diagonal elements of $\QXku$ are such that $\qu{x_i} = \sum_{x_j \ne x_i} \qu{x_i x_j}$, where $\qu{x_i}$ is the parameter of the exponential distribution associated with state $x_i$ of variable $X_k$. Therefore, $1/\qu{x_i}$ is the expected time that variable $X_k$ stays in state $x_i$ before transitioning to a different state $x_j$ when $\U = \u$. 
The off-diagonal elements $\qu{x_i x_j}$ are proportional to the probability that $X_k$ transitions from state $x_i$ to state $x_j$ when $\U = \u$.

Note that, conditional on $X_k$, $\QXku$ can be equivalently summarized with two independent sets of parameters: 
\begin{itemize}
    \item $\qXku = \left\{ \qu{x_i}, \forall x_i\in \Val(X_k) \right\}$, the set of intensities of the exponential distributions of the \textit{waits until the next transition}; and
    \item $\btXku = \left\{ \theta_{x_i x_j \given \u} = \qu{x_i x_j} / \qu{x_i}, \forall x_i, x_j \in \Val(X_k), x_i\neq x_j \right\}$, the \textit{probabilities of transitioning to specific states}.
\end{itemize}
Therefore, a CTBN $\N$ over $\X$ can be equivalently described by a graph $\G$ together with the corresponding sets of parameters
\begin{equation*}
\q = \left\{\qXku: \forall X_k \in \X, x_i \in \Val(X_k), \u \in \Val(\U) \right\}
\end{equation*}
and
\begin{equation*}
\boldsymbol{\Theta} = \left\{\btXku: \forall X_k \in \X, x_i \in \Val(X_k), \u \in \Val(\U)\right\}.
\end{equation*}
\hspace{1cm}

We assume that only one variable  in the CTBN can change state at any specific instant; and that its transition dynamics are specified by its parents via the CIM, while being independent of all other variables given its Markov Blanket.\footnote{The definition of Markov blankets in CTBNs is the same as in BNs: a Markov blanket comprises the parents, the children and the spouses of the target node; and it graphically separates the target node from the rest of the network.}

\begin{figure}[t]
\begin{center}
\includegraphics[width=1\linewidth]{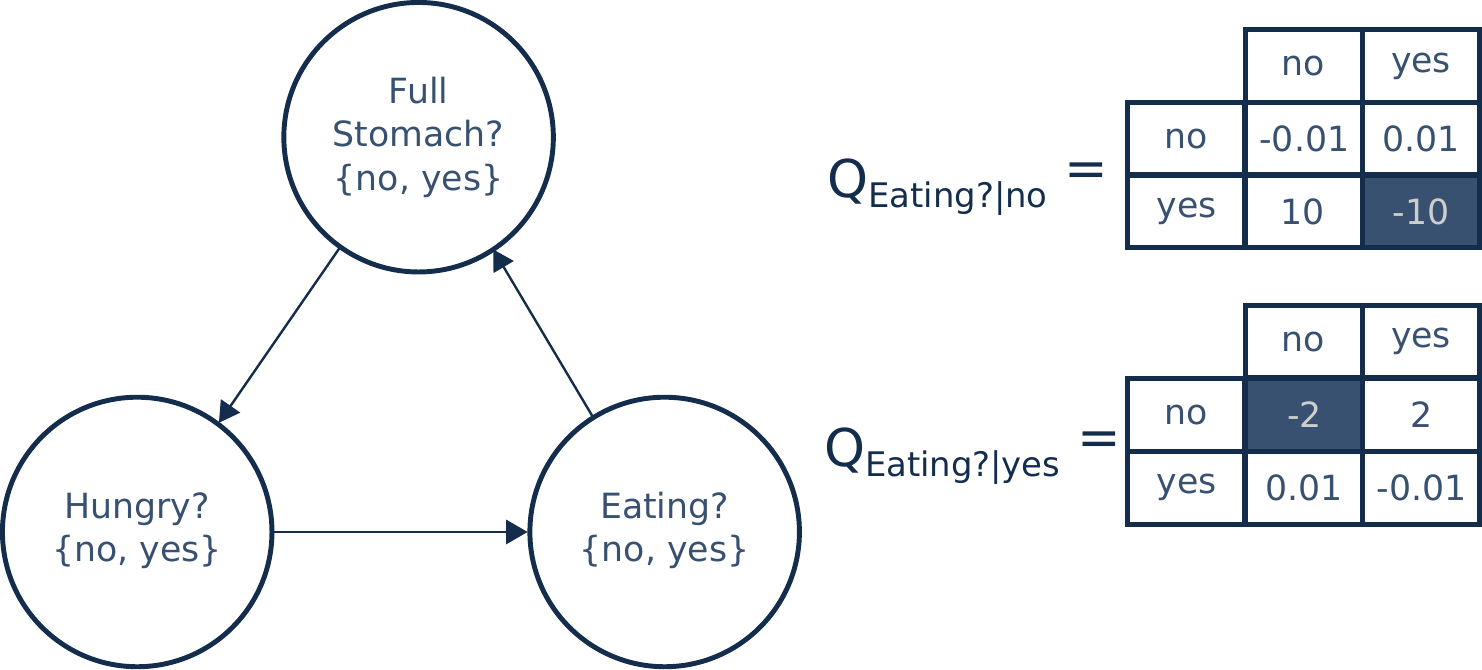}
\caption{The {\fontfamily{qcr}\selectfont Eating?}, {\fontfamily{qcr}\selectfont Full Stomach?}, {\fontfamily{qcr}\selectfont Hungry?} CTBN.}
\label{smallCTBN}
\end{center}
\end{figure}

Figure~\ref{smallCTBN} shows a CTBN which models the effect that eating ({\fontfamily{qcr}\selectfont Eating?}) has on the content of the stomach ({\fontfamily{qcr}\selectfont Full Stomach?}) of an individual. Furthermore, the content of the stomach has an effect on whether the individual is hungry or not ({\fontfamily{qcr}\selectfont Hungry?}), which is turn has an effect on the individual to start eating ({\fontfamily{qcr}\selectfont Eating?}).

The CTBN contains a cycle,
\begin{center}
{\fontfamily{qcr}\selectfont
{Eating?$ \hspace{0.2cm} \rightarrow$ \hspace{0.0cm} Full Stomach?$ \hspace{0.2cm} \rightarrow$ \hspace{0.0cm} Hungry?$ \hspace{0.2cm} \rightarrow$ \hspace{0.0cm} Eating?}}, 
\end{center}
which implies that whether a person is hungry or not ({\fontfamily{qcr}\selectfont Hungry?}) depends on whether the person's stomach is full or not ({\fontfamily{qcr}\selectfont Full Stomach?}), which depends on whether the person is eating or not ({\fontfamily{qcr}\selectfont Eating?}), which in turn depends on whether the person is hungry or not ({\fontfamily{qcr}\selectfont Hungry?}). 

Therefore, the CTBN consists of three nodes, that is, {\fontfamily{qcr}\selectfont Eating?}, {\fontfamily{qcr}\selectfont Full Stomach?} and {\fontfamily{qcr}\selectfont Hungry?}, associated with three binary random variables taking value on the sets $\Val$({\fontfamily{qcr}\selectfont Eating?}) = $\Val$({\fontfamily{qcr}\selectfont Full Stomach?}) = $\Val$({\fontfamily{qcr}\selectfont Hungry?}) = $\{no, yes\}$. 

The CIM for {\fontfamily{qcr}\selectfont Eating?} shown in Figure~\ref{smallCTBN} means that we
expect that a person who is hungry ({\fontfamily{qcr}\selectfont Hungry?}=$yes$) and not eating
({\fontfamily{qcr}\selectfont Eating?}=$no$), to begin eating, on average, in $1/2$ hours (-2 in
the filled cell), while we expect a person who is not hungry ({\fontfamily{qcr}\selectfont
Hungry?}=$no$) and who is eating ({\fontfamily{qcr}\selectfont Eating?}=$yes$), on average, to stop
eating in $1/10$ hours (-10 in the filled cell), that is, in 6 minutes. 
\footnote{{The CIMS for the other two nodes are available in Appendix}~\ref{appendix:extended_cims}.}

\subsection{Structure Learning}

Let $\D = \{\sigma_1, \ldots, \sigma_h\}$ be a sample consisting of $h$ trajectories denoted as $\sigma_j=\{\langle t_1, X_{t_1}\rangle, \ldots, \langle T_j, X_{T_j}\rangle\}$, where $T_j$ represents the length of trajectory $\sigma_j$, that is, the number of  transitions. For each pair $\langle t_i, X_{t_i}\rangle$, we denote the time of the $i$th transition as $t_i$ and the variable that leaves its current state at that time as $X_{t_i}$. 

Learning the structure of a CTBN from $\D$ can be cast as an optimization problem \citep{nsk03} in which we would like to find the graph $\G^*$ with the highest posterior log-probability given $\D$:
\begin{equation}
	\ln \Prob(\G \given \D) =  \ln \Prob(\G) + \ln \Prob(\D  \given  \G)
\label{Bayesian_score}
\end{equation}
where $\Prob(\G)$ is the \textit{prior distribution over the space of graphs} spanning $\X$ and $\Prob(\D  \given  \G)$ is the \textit{marginal likelihood} of the data given $\G$ averaged over all possible parameter sets.

The prior $\Prob(\G)$ is usually assumed to satisfy the \textit{structure modularity} property \citep{fk00}, so that it decomposes as
\begin{equation}\label{eq_str_mod}
	\Prob(\G) = \prod_{X_k \in \X} \Prob(Pa(X_k) = \U).
\end{equation} 
For simplicity, the literature often assumes a uniform prior, that is, $\Prob(\G) \propto 1$.

The marginal likelihood $\Prob(\D \given \G)$ depends on the parameter prior $\Prob(\q, \boldsymbol{\Theta} \given \G)$, which is usually assumed to satisfy the \textit{global parameter independence}, the \textit{local parameter independence} and the \textit{parameter modularity} properties \citep{heckerman} outlined below.
\begin{itemize}
    \item \textit{Global parameter independence}: the parameters $\qXkU$ and $\btXkU$ associated with each variable $X_k$ in a graph $\G$ are independent:
    \begin{equation} \label{eq_global_par_ind}
	    \Prob(\q, \boldsymbol{\Theta} \given \G) = \prod_{X_k \in \X} \Prob(\qXkU, \btXkU  \given \G).
    \end{equation} 
    \item \textit{Local parameter independence}: for each variable $X_k$, the parameters associated with each configuration $\u$ of parent set $\U$ are independent:
    \begin{equation}\label{eq_local_par_ind}
	    \Prob(\qXkU, \btXkU  \given \G) = 
	    \prod_{\u \in \Val(\U)} \; \prod_{x_i \in \Val(X_k)} \Prob(\q_{x_i \given \u},\btheta_{x_i \given \u} \given \G).
    \end{equation}
    \item \textit{Parameter modularity}: if variable $X_k$ has the same parent set in two distinct graphs $\G$ and $\G'$, then the prior probability for the parameters associated with $X_k$ should also be the same:
    \begin{equation}\label{eq_par_mod}
        \Prob(\qXkU, \btXkU \given \G) = \Prob(\qXkU, \btXkU \given \G') .
    \end{equation}
\end{itemize}
In the context of CTBNs, we assume that the priors over the waiting times and over the transition probabilities are independent as well:
\begin{equation}\label{eq_par_indep_exp_mult}
	\Prob(\q, \boldsymbol{\Theta} \given \G) = \Prob(\q \given \G) \Prob(\boldsymbol{\Theta} \given \G). 
\end{equation} 
\citet{nsk03} suggested conjugate priors for both $\q$ and $\boldsymbol{\Theta}$ in the form of
\begin{align}
	\Prob(\q_{x_i \given \u}) &\sim
	  \Ga \left( \aiu, \tau_{x_i \given \u} \right),  \label{eq_prior_gam} \\ \label{eq_prior_dir}
	\Prob(\btheta_{x_i \given \u}) &\sim
	  \Dir \left(\alpha_{x_i x_1 \given \u}, \dots, \alpha_{x_i x_m \given \u}\right),
\end{align}
where $\aiu, \tiu, \alpha_{x_i x_1 \given \u}, \dots, \alpha_{x_i x_m \given \u}$ are the priors' hyperparameters.

In particular, for any $X_k \given \U = \u$, $\aiu$ and $\aiju$ represent the pseudocounts for the number of transitions from state $x_i$ to state $x_j$; and $\tiu$ represents the imaginary amount of time spent in each state $x_i$ before any data is observed.
Note that $\aiu$ is inversely proportional to the number of joint states of the parents of $X_i$.
After conditioning on  the dataset $\D$, we obtain the following posterior distributions:
\begin{align}
	\Prob(\q_{x_i \given \u} \given \D) &\sim  
	  \Ga \left( \aiu + M_{x_i \given \u}, \tau_{x_i \given \u} + T_{x_i \given \u} \right),   \label{eq_par_cond1} \\
	\Prob( \btheta_{x_i \given \u} \given \D ) &\sim 
      \Dir \left(\alpha_{x_i x_1 \given \u}+ M_{x_i x_1 \given \u}, \dots, \alpha_{x_i x_m \given \u} + M_{x_i x_m \given \u} \right), \label{eq_par_cond2}
\end{align}
where $T_{x_i \given \u}$ and $M_{x_i x_j \given \u}$ are the sufficient statistics of the CTBN. 

In particular, $T_{x_i \given \u}$ is the amount of time spent by $X_k$ in the state $x_i$ and $M_{x_i x_j \given \u}$ is the number of times that $X_k$ transitions from the state $x_i$ to the state $x_j$, given $\U = \u$.\footnote{The number of times $X_k$ leaves the state $x_i$ when $\U = \u$ is $M_{x_i \given \u} = \sum_{x_j \ne x_i} M_{x_i x_j  \given \u}$.} 

The marginal likelihood $\Prob(\D \given \G)$ arising from these posteriors can be written as
\begin{equation}\label{eq_ctbn_mlike}
	\Prob (\D  \given  \G )  =  \prod_{X_k \in \X} \ML(\qXkU : \D)  \ML(\btXkU : \D)
\end{equation}
due to (\ref{eq_global_par_ind}) and (\ref{eq_par_indep_exp_mult}).
$\ML(\qXkU : \D)$ is the marginal likelihood of $\qXkU$,
\begin{equation}\label{eq_ctbn_ml_q}
	 \ML(\qXkU : \D) = \prod_{\u \in \Val(\U)} \prod_{x_i\in \Val(X_k)} \frac{\Gamma \left(\aiu + M_{x_i \given \u}+1\right) \left( \tau_{x_i \given \u} \right)^{\aiu + 1} }{\Gamma \left( \aiu + 1 \right) \left( \tau_{x_i \given \u} + T_{x_i \given \u} \right)^{\aiu + M_{x_i \given \u}+1}};
\end{equation}
and $\ML(\btXkU : \D)$ is the marginal likelihood of $\btXkU$,
\begin{equation}\label{eq_ctbn_ml_theta}
	 \ML(\btXkU : \D) = \prod_{\u\in \Val(\U)} \prod_{x_i\in \Val(X_k)} \frac{\Gamma \left(\aiu\right)}{\Gamma\left(\aiu + M_{x_i \given \u} \right)} 
\prod_{\stackrel{x_j\in \Val(X_k)}{}} \frac{\Gamma\left(\aiju + M_{x_i x_j \given \u} \right)}{\Gamma \left(\aiju\right)}.
\end{equation}
The resulting $\Prob(\D \given \G)$ is the Bayesian-Dirichlet equivalent (BDe) metric for CTBNs \citep{n07} based on the priors
(\ref{eq_prior_gam}) and (\ref{eq_prior_dir}), which satisfies assumptions (\ref{eq_global_par_ind}), (\ref{eq_local_par_ind}), and (\ref{eq_par_mod}) by construction. 

The posterior in (\ref{Bayesian_score}) can then be written in closed form as 
\begin{equation}
	\Prob(\G \given \D)  =   \sum_{X_k \in \X}  \log \Prob(Pa(X_k)=\U) + \log \ML(\qXkU : \D) + \log \ML(\btXkU : \D)
 	\label{eq_ctbn_bs_final}
\end{equation}
assuming that (\ref{eq_str_mod}) is satisfied.
Since $\G$ does not have acyclicity constraints in a CTBN, it is possible to maximize (\ref{eq_ctbn_bs_final}) by independently scoring the possible parent sets of each $X_k$. Therefore, if we bound the maximum number of parents we can find the optimal $\Prob(\G \given \D)$ in polynomial time either by enumerating all possible parent sets or by using hill-climbing to add, delete or reverse arcs \citep{nsk03}. The author didn't give a proper name to this algorithm. However we decided to call it \textbf{C}ontinous-\textbf{T}ime \textbf{S}earch and \textbf{S}core (CTSS).

\section{A Constraint-Based Algorithm for Structure Learning}

Learning the structure of a BN is a problem that is well explored in the literature. Several approaches have been proposed spanning score-based, constraint-based and hybrid algorithms; recent reviews are available from \citet{ijar19,Scanagatta2019ASO}. 
Score-based algorithms find the BN structure that maximizes a given score function, while constraint-based algorithms use statistical tests to learn conditional independence relationships (called \textit{constraints}) from the data and infer the presence or absence of particular arcs. Hybrid algorithms combine aspects of both score-based and constraint-based algorithms.

On the other hand, the only structure learning algorithm proposed for CTBNs is the CTSS algorithm
from \citet{nsk03} we described in the previous section: to the best of our knowledge no
constraint-based algorithm exists in the literature. 
{It is worthwhile to note that CTBNs are a special case of the local independence model
\citep{didelez2008graphical,didelez2012asymmetric,meek2014toward} for which a
general structure learning algorithm has been made available. In particular,
\mbox{\citep{mogensen2018causal}} proposed a structure learning algorithm for local independence
graphs. In their work, the authors studied independence models induced by directed graphs (DGs) by
formalizing the properties of abstract graphoids that ensure that the global Markov property holds
for a given directed graph. \mbox{\citet{mogensen2018causal}} applied their theoretical arguments to the Ito diffusion as
well as the event process, which are both related to CTBNs. Therefore, the algorithm we propose here can
be considered as an instance of that presented by \mbox{\citet{mogensen2018causal}}, even if we independently formulated
and developed our structure learning algorithm that is specifically designed
for CTBNs.}
After a brief introduction to constraint-based
algorithms for BNs, we propose such an algorithm for CTBNs.

\subsection{Constraint-Based Algorithms for BNs}
\label{sec:bnsl}

Constraint-based algorithms for BN structure learning originate from the \textit{Inductive Causation} (IC) algorithm from \citet{PeVe91} for learning causal networks.
IC starts (step 1) by finding pairs of nodes connected by an undirected arc as those are not independent given any other subset of variables. The second step (step 2) identifies the v-structures $X_i \rightarrow X_k \leftarrow X_j$ among all pairs $X_i$ and $X_j$ of non-adjacent nodes which share a common neighbour $X_k$. Finally, step 3 and 4 of IC identify compelled arcs and orient them to build the completed partially oriented DAG (CPDAG) that describes the \textit{equivalence class} the BN falls into.

\begin{algorithm}[t]
    \begin{enumerate}
        \item Form the complete undirected graph $\G$ on the vertex set $\X$.
        \item For each pair of variables $X_i,X_j \in \X$, consider all the possible 
            separating set from the smallest ($\Sij = \varnothing$) to the largest ($\Sij = \X \setminus \{X_i, X_j\}$). If there isn't any set $\Sij$ such that $X_i \indep X_j \given \Sij$ then the edge $X_i \uar X_j$ is removed from $\G$.
        \item For each triple $X_i,X_j,X_k \in \G$ such that $X_i \uar X_j$, 
            $X_j \uar X_k$, and $X_i$, $X_j$ are not connected, orient the edges into $X_i \rightarrow X_j \leftarrow X_k$ if and only if $X_j \not\in \Sij$ for every $\Sij$ that makes $X_i$ and $X_k$ independent.
        \item The algorithm identifies the compelled directed arcs by iteratively applying the following two rules:
        \begin{enumerate}
      \item if $X_i$ is adjacent to $X_j$ and there is a strictly directed path
        from $X_i$ to $X_j$ then replace $X_i \uar X_j$ with $X_i \rightarrow X_j$
        (to avoid introducing cycles);
      \item if $X_i$ and $X_j$ are not adjacent but $X_i \rightarrow X_k$ and
        $X_k \uar X_j$, then replace the latter with $X_k \rightarrow X_j$ (to
        avoid introducing new v-structures).
        \end{enumerate}
        \item Return the resulting CPDAG $\G$.
    \end{enumerate}
    \caption{PC Algorithm}
    \label{PC-algo}
\end{algorithm}

However, steps 1 and 2 of the IC algorithm are computationally unfeasible for non-trivial problems due to the exponential number of conditional independence relationships to be tested. 

The \textit{PC algorithm}, which is briefly illustrated in Algorithm~\ref{PC-algo}, was the first proposal addressing this issue; its modern incarnation is described in \citet{colombo}, and we will use it as the foundation for CTBN structure learning below. 
PC starts from a fully-connected undirected graph. Then, for each pair of variables $X_i$, $X_j$ it proceeds by gradually increasing the cardinality of the set of conditioning nodes $\Sij$ until $X_i$ and $X_j$ are found to be independent or $\Sij = \X \setminus \{X_i, X_j\}$. The remaining steps are identical to those of IC.

Neither IC nor PC (or other constraint-based algorithms, for that matter) require a specific test statistic to test conditional independence, making them independent from the distributional assumptions we make on the data.

\subsection{The CTPC Structure Learning Algorithm}
\label{sec:proposal}

CTBNs differ from BNs in three fundamental ways: BNs do not model time, while CTBNs do; BNs are based on DAGs, while CTBNs allow cycles; and BNs model the dependence of a node on its parents using a conditional probability distribution, while CTBNs model it using a CIM. These differences make structure learning a simpler problem for CTBNs than it is for BNs.

Firstly, learning arc directions is an issue in BNs but not in CTBNs, where arcs are required to follow the arrow of time. Unlike BNs, which can be grouped into equivalence classes that are probabilistically indistinguishable, each CTBN has a unique minimal graphical representation \citep{nsk03}. 
For instance, let a CTBN $\N$ have graph $\G = \{X\rightarrow Y\}$: unless trivially $X$ and $Y$ are marginally independent, $\G$ cannot generate the same transition probabilities as any CTBN $\N'$ with graph $\G' = \{X\leftarrow Y\}$.

Secondly, in CTBNs we can learn each parent set $Pa(X_k)$ in isolation, thus making any structure learning algorithm embarrassingly parallel. Acyclicity imposes a global constraint on $\G$ that makes it impossible to do the same in BNs.

Thirdly, each variable $X_k$ is modelled conditional on a given function of its parent set $Pa(X_k)$: a conditional probability table for (discrete) BNs, a CIM for CTBNs.

However, a CIM $\QXkU$ describes the temporal evolution of the state of variable $X_k$ conditionally on the state of its parent set $\U$. Hence we can not test conditional independence by using  classical test statistics like the mutual information or Pearson's $\chi^2$ that assume observations are independent \citep{koller}.
Instead we need to adapt our definition of conditional independence to CTBNs in order to design a constraint-based algorithm for structure learning.

\begin{definition} Conditional Independence in a CTBN

Let $\N$ be a CTBN with graph $\G$ over a set of variables $\X$. We say that $X_i$ is conditionally independent from $X_j$ given $\Sij \subseteq \X \setminus \{X_i, X_j\}$ if
\begin{align}
    &\Q_{{X}_i \given x, \s} = \Q_{{X}_i \given  \s}&
    &\forall\, x\in \Val(X_j), \forall\, \mathbf {s} \in \Val(\Sij).
    \label{CTBN_cond_indep_equa}
\end{align}
If $\Sij = \varnothing$, then $X_i$ is said to be marginally independent from $X_j$.

\label{CTBN_cond_indep_def}
\end{definition}

\noindent It is important to note that Definition~\ref{CTBN_cond_indep_def} is not symmetric: it is perfectly possible for $X_i$ to be conditionally or marginally independent from $X_j$, while $X_j$ is not conditionally or marginally independent from $X_i$. This discrepancy is, however, not a practical or theoretical concern because arcs are already non-symmetric (they must follow the direction of time) and therefore we only test whether $X_i$ depends on $X_j$ if $X_j$ precedes $X_i$ and not the other way round.

As for the test statistics, we can test for conditional independence using $\qXku$ (the waiting times) and, if we do not reject the null hypothesis of conditional independence, we can perform a further test using $\btXku$ (the transitions): $\qXku$ and $\btXku$ have been defined to be independent in Section~\ref{sec:notation} so they can be tested separately.

Note that conditional independence can be established by testing only the waiting times $\qXku$ if the CTBN contains only binary variables because the transition is deterministic for a binary node. 
However, testing for conditional independence involves both waiting times and transitions in the general case in which variables can take more than two values. 

Since we consider that rates are the most important characteristic to  assess in a stochastic process, we decide without loss of generality to test $\qXku$ first, and then $\btXku$.

For $\qXku$, we define the \textit{null time to transition hypothesis} as follows.

\begin{definition} Null Time To Transition Hypothesis

Given $X_i$, $X_j$ and the conditioning set $\Sij \subseteq \X \setminus \{X_i, X_j\}$, the null time to transition hypothesis of $X_j$ over $X_i$ is
\begin{align*}
    &q_{x \given y,\s} = q_{x \given \s}&
    &\forall\, x \in \Val(X_i), \forall\, y \in \Val(X_j), \forall\, \s \in \Val(\Sij).
\end{align*}
\label{NTTTH}
\end{definition}

For $\btXku$, we define the \textit{null state-to-state transition hypothesis} as follows.

\begin{definition} Null State-To-State Transition Hypothesis

Given $X_i$, $X_j$ and the conditioning set $\Sij \subseteq \X \setminus \{X_i, X_j\}$, the null state-to-state transition hypothesis of $X_j$ over $X_i$ is
\begin{align*}
    &\theta_{x\cdot \given y,\s} = \theta_{x\cdot \given \s}& 
    &\forall\, x \in \Val(X_i),  \forall\, y \in \Val(X_j), \forall\, \s \in \Val(\Sij)
\end{align*}
where we let $\theta_{x\cdot \given y,\s}$ be off diagonal elements of matrix $\Q_{X_i \given y,\s}$ divided by $q_{x \given y,\s}$ corresponding to assignment $X_i=x$. It is worthwhile to mention that equality $\theta_{x\cdot \given y,\s} = \theta_{x\cdot \given \s}$ has to be understood in terms of corresponding components of vectors $\theta_{x\cdot \given y,\s}$ and $\theta_{x\cdot \given \s}$.
\label{STST}
\end{definition}

Definition~\ref{NTTTH} characterizes conditional independence for the times to transition for variable $X_i$ when adding (or not) $X_j$ to its parents; Definition~\ref{STST} characterizes conditional independence for the transitions of $X_i$  when adding (or not) $X_j$ to its parents.

To test the \textit{null time to transition hypothesis}, we use the $F$ test to compare two exponential distributions from \citet{lee2003statistical}. In the case of CTBNs, the test statistic and the degrees of freedom take form
\begin{align}
    &F_{r_1,r_2} = \frac{q_{x \given \s}}{q_{x \given y, \s}},&
    &\text{with}&
    &r_1 =  \sum_{x'\in \Val(X_i)} \Mys,&  
    r_2 =  \sum_{x'\in \Val(X_i)} \Ms.
    \label{T2EM}
\end{align}

To test the \textit{null state-to-state transition hypothesis}, we investigated the use of the \textit{two-sample chi-square} and \textit{Kolmogorov-Smirnov} tests \citep{mitchell}. For CTBNs the former takes form:
\begin{align}
    &\chi ^2 = \sum_{x'\in \Val(X_i)} \frac{(K\cdot \Mys - L\cdot \Ms)^2}{\Mys + \Ms},&\\ \nonumber \\
    &K = \sqrt{\frac{\sum_{x'\in \Val(X_i)} \Ms}{\sum _{\stackrel{x'\in \Val(X_i)}{}}  \Mys}},
     L = \frac{1}{K}\;,
    \label{2SCS}
\end{align}
and is asymptotically distributed as a $\chi^2_{|\Val(X_i)|-1}$.
The latter is defined as
\begin{align}
    &D_{r_1,r_2} = \sup _{x' \in \Val(X_i)} \left| \Theta_{x x' \given \s} - \Theta_{x x' \given y, \s}\right|,
    &\Theta _{x x'} = \sum _{\stackrel{x'' \in \Val(X_i)}{x'' \leq x'}} \theta _{x x''}.
    \label{2SKS}
\end{align}

\begin{algorithm}[t!]
\renewcommand{\labelenumii}{\theenumii}
\renewcommand{\theenumii}{\theenumi.\arabic{enumii}}
\renewcommand{\labelenumiii}{\theenumiii}
\renewcommand{\theenumiii}{\theenumii.\arabic{enumiii}}
\makeatletter
\renewcommand\p@enumiii{}
\makeatother
\begin{enumerate}
  \item Form the complete directed graph $\G$ on the vertex set $\X$. \label{algo-init-G}
  \item For each variable $X_i \in \X$: \label{algo-iteratie-X}
  \begin{enumerate}
    \item Set $\U = \{X_j \in \X: X_j \rightarrow X_i\}$, the current parent set. \label{algo-tested-variables}
    \item For increasing values $b = 0, \ldots, n$, until $b = |\U|$:
    \begin{enumerate}
      \item For each $X_j \in \U$, test $X_i \indep X_j \given \Sij$ for all possible subsets of size $b$ of 
        $\U \setminus X_j$. \label{algo-testing}
      \item As soon as $X_i \indep X_j \given \Sij$ for some $\Sij$, remove $X_j \rightarrow X_i$ from $\G$
        and $X_j$ from $\U$. \label{algo-remove-edge}
    \end{enumerate}
  \end{enumerate}
  \item Return directed graph $\G$.
\end{enumerate}
\caption{Continuous-time PC Algorithm}
\label{CTPC}
\end{algorithm}

After characterising conditional independence, we can now introduce our constraint-based algorithm for structure learning in CTBNs. The algorithm, which we call \textit{Continuous-Time PC} (CTPC), is shown in Algorithm~\ref{CTPC}.

The first step is the same as the corresponding step of the PC algorithm in that it determines the same pattern of conditional independence tests. However, as discussed above, the hypotheses being tested are the \textit{null time to transition hypothesis} and the \textit{null state-to-state transition hypothesis}.

The second step of CTPC differs from that in the PC algorithm. Since independence relationships are not symmetric in CTBNs, we can find the graph $\G$ of the CTBN without indentifying and then refining a CPDAG representing an equivalence class. Therefore, steps 3 and 4 of the PC algorithm are not needed in the case of CTBNs.

CTPC starts by initializing the complete directed graph $\G$ without loops (step~\ref{algo-init-G}).  Note that while loops (that is, arcs like $X_i \rightarrow X_i$) are not included, cycles of length two (that is, $X_i \rightarrow X_j$ and $X_j \rightarrow X_i$) are, as well as cycles of length tree or more.

Step~\ref{algo-iteratie-X} iterates over the $X_i$ to identify their parents $\U$. This is achieved in step~\ref{algo-testing} by first testing for unconditional independence, then by testing for conditional independence gradually increasing the cardinality $b$ of the considered separating sets. 

Each time Algorithm~\ref{CTPC} concludes that $X_i$ is independent from $X_j$ given some separating set, we remove the arc from node $X_j$ to node $X_i$ in step~\ref{algo-remove-edge}. At the same time, we also remove $X_j$ from the current parent set $\U$. The iteration for $X_i, X_j$ terminates either when $X_j$ is found to be independent from $X_i$ or when there are no more larger separating sets to try because $b = |U|$; and the iteration over $X_i$ terminates when there are no more $X_j$ to test.

CTPC checks the \textit{null time to transition hypothesis} (Definition~\ref{NTTTH}) by applying the \textit{test for two exponential means} in (\ref{T2EM}). On the contrary, the \textit{null state-to-state transition hypothesis} (Definition~\ref{STST}) can be tested using two different tests: the \textit{two sample chi-square test} in (\ref{2SCS}) and the \textit{two sample Kolmogorov-Smirnov test} in (\ref{2SKS}). We call these two options CTPC\textsubscript{$\chi^2$} and CTPC\textsubscript{KS}, respectively.

{The proposed algorithm is able to recover the true graph under the standard assumptions of the PC
algorithm: {\em i}) the faithfulness assumption; {\em ii}) the database consists of a set of independent and identically distributed cases; {\em iii}) the database of cases is infinitely large; {\em iv}) the causal
sufficiency assumption, that is, no hidden (latent) variables are involved; {\em v}) the statistical tests make no type-I or type-II errors \mbox{\citep{KjMa13}}. 

CTPC can be also extended to account for hidden variables. In the case of score-based structure learning, \mbox{\citet{nodelman2012expectation}} used the Structural Expectation Maximization algorithm for this purpose, while
\mbox{\citet{LinznerSK19}} developed a novel gradient-based approach to structure learning which makes it possible to learn structures of previously inaccessible sizes. However, to the best of our knowledge no constraint-based algorithm for learning the structure of CTBNs with latent variables has been presented in the literature. The results presented and discussed in \mbox{\citet{mogensen2018causal}} may allow the CTPC to be extended to handle hidden variables.}

\subsection{CTPC computational complexity}
\label{sec:bigO}
The computational complexity of estimating the network structure of a CTBN from data using the CTPC
algorithm depends on the following quantities: the number of nodes, the number of parents of each
node and the number of transitions. 

In Appendix~\ref{appendix:complexity} we presented the computational complexity of
learning the stucture of a CTBN using CTPC summarized in the following table:

\begin{center}
    \begin{tabular}{c|c}
      \textbf{Case} & \textbf{Complexity}  \\
      \hline
      Best Case     & $O(n^2 \cdot (2\cdot\gamma ^3 + 2\cdot \psi))$ \\
      Worst case    & $O(n \cdot 2^{n} \cdot (\gamma^{n} + n \cdot \psi))$ \\
      General case  & $ O\left(\frac{n^{\rho + 2}}{(\rho + 1)!} \cdot (\gamma ^{\rho + 3} + (\rho + 2)  \cdot \psi )\right)$
    \end{tabular}
\end{center}

\noindent where $n$ is the number of nodes, $\gamma$ is the maximum node cardinality present in the network,
$\psi$ is the number of transitions occurring in the dataset and $\rho$ is the maximum parent set
size presente in the network.

\citet{n07} states that the CTSS algorithm for CTBNs is polynomial in the number of variables ($n$) and in the size of the dataset ($\psi$) when a maximum number of parents is given.
The CTPC algorithm is also polynomial in the number of nodes and the size of the dataset in the general case. However, in the CTPC algorithm it is natural to bound the size of the separating sets in the conditional independence tests but not the size of the parent sets because CTPC does not operate on parent sets explicitly. Bounding the size of the separating sets can reduce the execution time, but has the drawback of producing denser networks because it potentially makes some independence relationships impossible to establish.

\section{Numerical Experiments}
\label{sec:simulations}

We now assess the performance of CTPC against that of the CTSS algorithm from \citet{nsk03} using
synthetic data. In particular, we generate random CTBNs as the combinations of directed graphs and
the associated CIMs; and we generate random trajectories from each CTBN. 

Note that we only generate
connected networks, hence absolute density\footnote{absolute density = Number of edges in a network} is bounded below
by $n - 1$. 

We measure the performance of the learning algorithms using the F1 score over the arcs, which is defined as
\begin{align*}
  F_1 = 2 \cdot \frac{\mathrm{precision} \times \mathrm{recall}}{\mathrm{precision} + \mathrm{recall}}\;.
\end{align*}
Since there is no score equivalence in CTBNs, nor are networks constrained to be acyclic, comparing graphs is equivalent to evaluating a binary classification problem.

{Furthermore, we compare the two algorithms by their scaled difference in Bayesian
Information Criterion (BIC), the latter defined as in \mbox{\citet{koller}}}:
\begin{align*}
  \Delta BIC\% = \frac{BIC_{CTSS}-BIC_{CTCP}}{BIC_{CTSS}} \cdot 100
\end{align*}
\noindent with 
\begin{equation}
    BIC =\ln (\hat{L})  - \frac{1}{2} k \ln (\psi)
\end{equation}
{ and where $k$ is the number of parameters in the learned CTBN model, $\psi$ is the number of transitions while $\hat{L}$ is
the corresponding data likelihood.}

After a first set of experiments (Appendix~\ref{appendix:experiments1}) we found that the
CTPC\textsubscript{$\chi^2$} performs marginally better than CTPC\textsubscript{$KS$}. 
Then we set up a full factorial experimental design over different numbers of nodes $n = \{5, 10, 15, 20\}$, network densities\footnote{$\text{network density} = \frac{\text{absolute density}}{n \cdot (n-1)}$} $\{0.1, 0.2, 0.3, 0.4\}$, number of states for the nodes $|\Val(X_i)| = \{2, 3, 4\}$. For
each network, we generate 300 trajectories that last on average 100 units of time each. We perform
10 replicates for each simulation configuration except for the networks with $20$ ternary nodes and
network density equal to $0.4$, for which we only perform 3 replicates. We did not consider quaternary
networks with 20 nodes because it is unfeasible to learn them on the available hardware. 

We perform the experiments using new, optimized implementations of the CTSS algorithm and of the
CTPC\textsubscript{$\chi^2$} algorithm\footnote{The new implementations provided by Filippo Martini
and Luca Moretti are available at \url{https://github.com/madlabunimib/PyCTBN}.} that can handle
larger networks and that can learn the parent set of each node in parallel. Those implementations
are parallelized but use more memory, requiring a more powerful machine with 8 cores and 64GB of
memory.

\begin{figure}[H]
    \centering
    \includegraphics[width=1.02\textwidth]{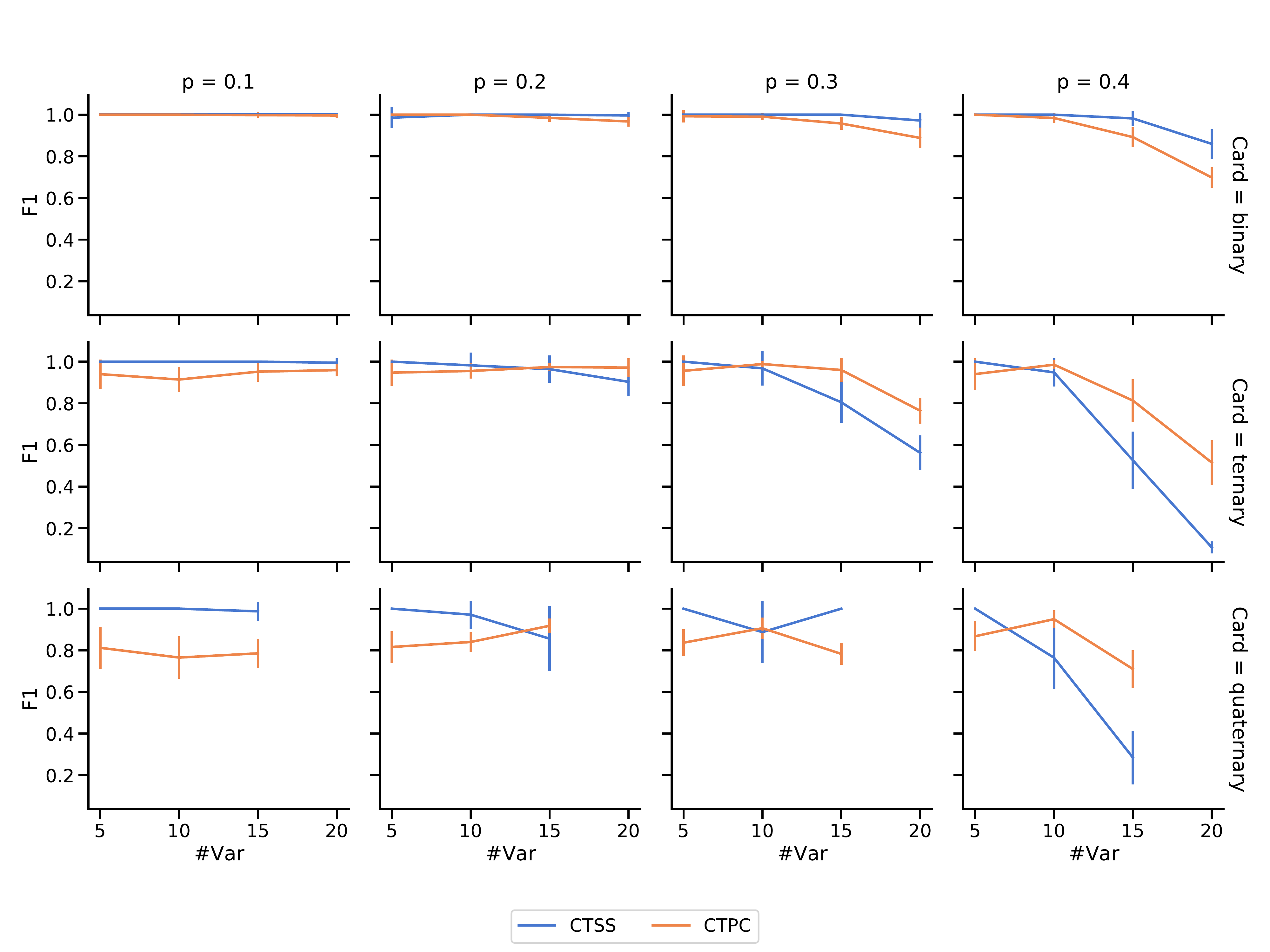}
    \caption{Each of the plots on this figure represents the average F1 score and the standard deviation of the constraint-based algorithm and the CTSS one, against the number of nodes for a specific combination of network density and node cardinality.}
    \label{fig:lineplot_f1}
\end{figure}

\begin{figure}[H]
    \centering
    \includegraphics[width=\textwidth]{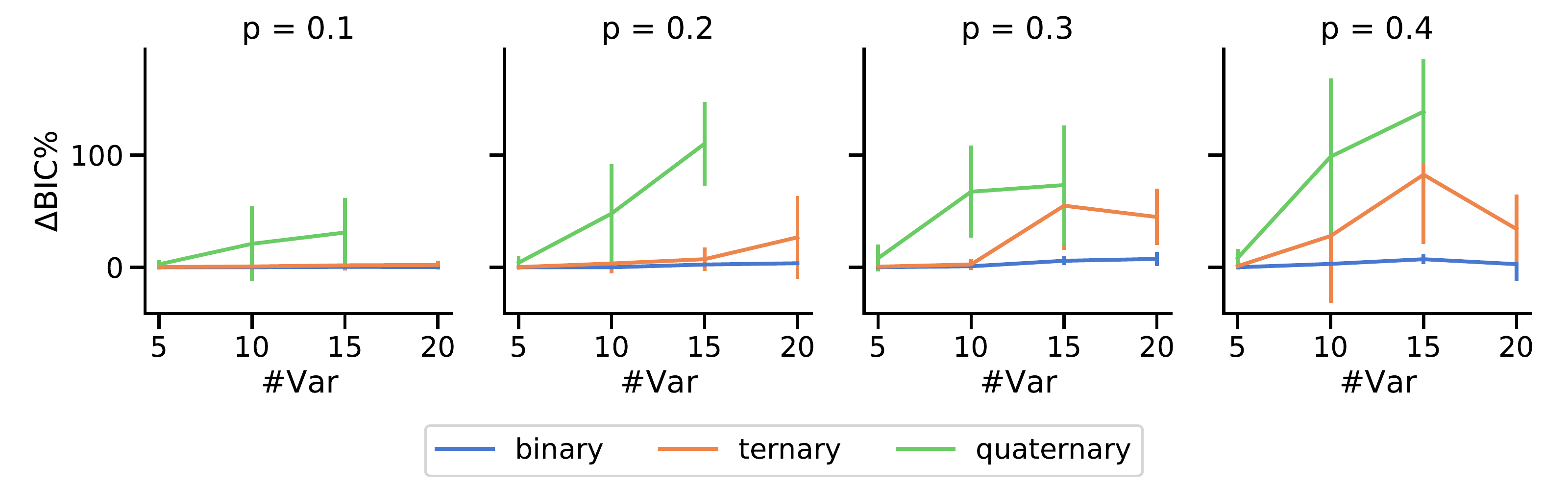}
    \caption{Each of the plots on this figure represents the average $\Delta BIC \%$  and the standard deviation against the number of nodes for a specific combination of network density and node cardinality.}
    \label{fig:lineplot_deltabic}
\end{figure}

\begin{figure}[b!]
    \centering
    \includegraphics[width=1.02\textwidth]{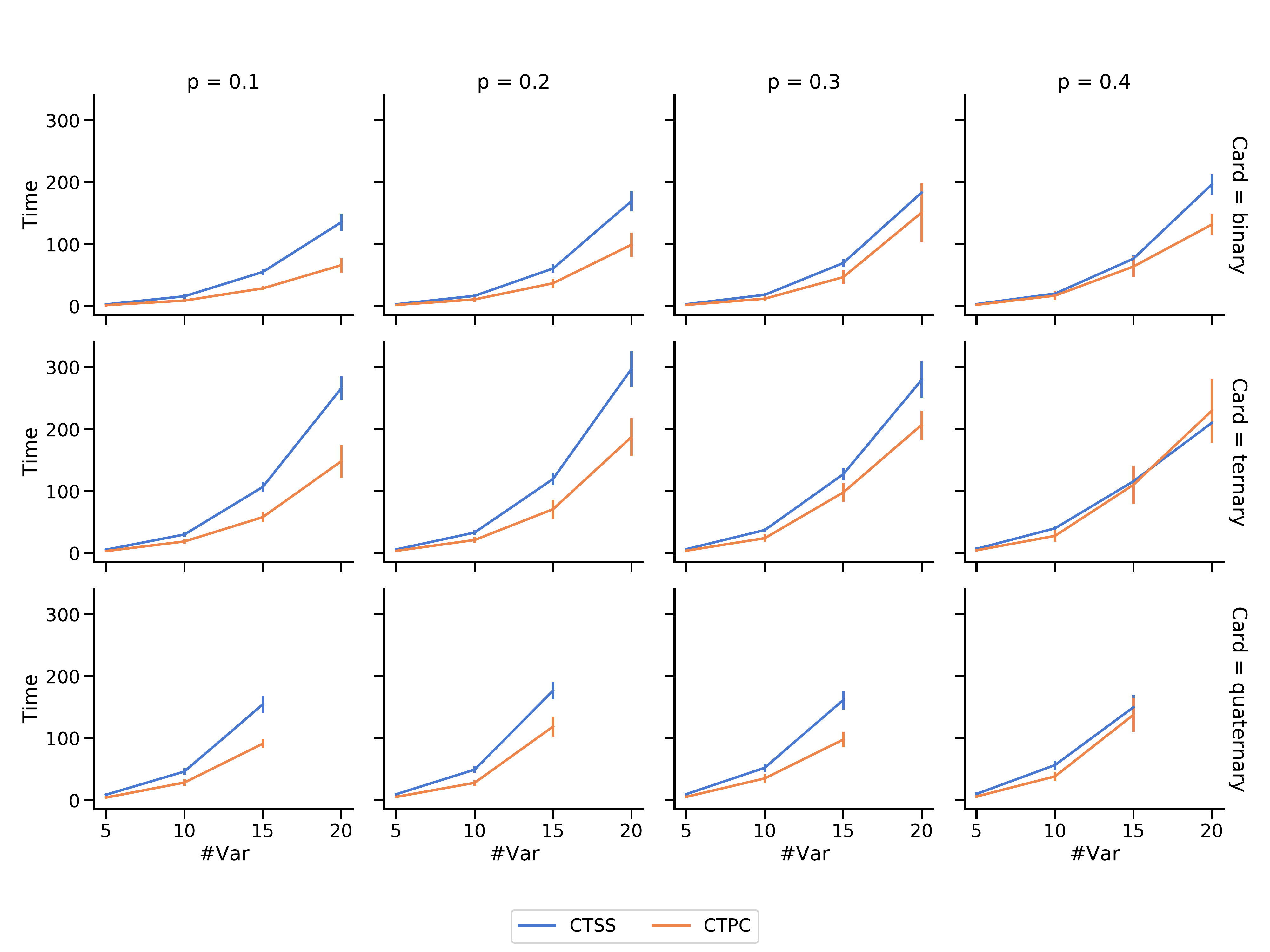}
    \caption{Each of the plots on this figure represents the average execution time in seconds and the standard deviation of the constraint-based algorithm and the CTSS one, against the number of nodes for a specific combination of network density and node cardinality.}
    \label{fig:lineplot_time}
\end{figure}

\begin{figure}[H]
    \centering
    \includegraphics[width=1.02\textwidth]{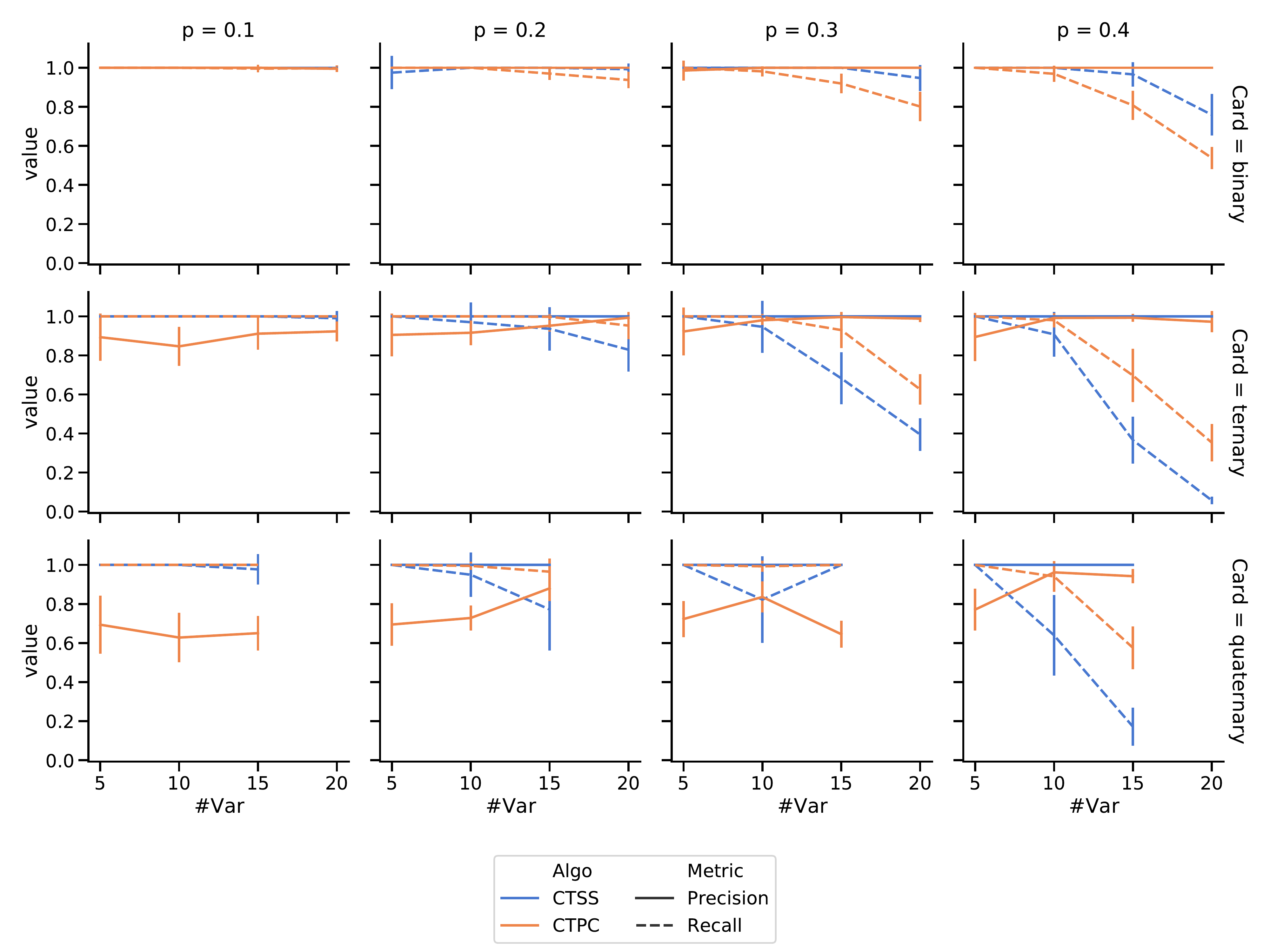}
    \caption{Each of the plots on this figure represents the average Precision, Recall and their respective standard deviations of the constraint-based algorithm and the CTSS one, against the number of nodes for a specific combination of network density and node cardinality.}
    \label{fig:lineplot_pr}
\end{figure}

The results of our simulation study are summarized in Figure~\ref{fig:lineplot_f1}, in Figure \ref{fig:lineplot_deltabic}, in Figure~\ref{fig:lineplot_time} and in Figure~\ref{fig:lineplot_pr}. 

Figure~\ref{fig:lineplot_f1} shows that the CTSS algorithm is the best choice for networks consisting of binary nodes. However, the CTPC\textsubscript{$\chi^2$} and the CTSS perform similarly for networks with ternary and quaternary nodes. Furthermore, the performance of the CTPC algorithm, if compared with the CTSS one, seems to improve with the increase in the number of nodes and in the network density. 

We can see from Figure~\ref{fig:lineplot_f1} and Figure~\ref{fig:lineplot_time} that the execution time increases when the network density increases, and that both algorithms perform poorly for dense networks. This behaviour may be attributed to the method used to generate the trajectories that doesn't increase their size when the network density increases.

{Figure \mbox{\ref{fig:lineplot_deltabic}} shows something different. Indeed, if we consider the $\Delta BIC\%$ as an evaluation metric, we note the CTSS algorithm to be at least as good as the CTPC algorithm in almost every experiment.}

Figure~\ref{fig:lineplot_pr} shows an important difference between the two algorithms; the CTSS algorithm achieves a precision of 1 in almost all simulations. However, the constraint-based algorithm has a better recall for the networks with ternary and quaternary nodes\footnote{The tables with all the results can be found in the Appendix \ref {appendix:experiments2}}. 

{Until now, we tested the CTPC algorithm with small to medium networks consisting of up to 20 nodes. To evaluate the execution time of CTPC with larger networks, we combined up to 5 networks with 20 binary nodes and network density $0.1$.  The results presented in Table
\mbox{\ref{table:bigNetworks}}
show that CTPC is able to learn networks with up to 100 nodes in a reasonable amount of time.
It is important to note that the tested networks are \textit{scattered}.~\footnote{Networks composed by disjoint sub-networks that are collated together in such a way to maintain
sparsity} It was also necessary to use a
machine with 256GB of memory to carry out the experiments. The huge memory requirement is closely related to our implementation. In fact, each process is independent and requires a copy of the dataset to be loaded into the ram.}
\begin{table}[t]
  \centering
\begin{tabular}{ccc}
  Cardinality & Execution Time (single core) & Execution Time (two cores)\\ \hline
  20  & 4m & 4m\\
  40  & 50m & 50m\\
  60  & 4h15m & 3h40\\
  80  & 13h30m & 8h47\\
  100 & 30h28m & 21h19m \tablefootnote{This value was estimated as it was not possible to perform the experiment with the amount of memory available to us.}
\end{tabular}
\caption{Execution time of the CTPC on big, sparse networks}
\label{table:bigNetworks}
\end{table}

\section{Conclusions}
\label{sec:conclusions}

In this paper we introduced the first constraint-based algorithm for structure learning in CTBNs, which we called CTPC, comprising both a suitable set of statistics for testing conditional independence and a heuristic algorithm based on PC. We also derived the
complexity of this new algorithm finding that it is similar to the CTSS one. 

CTPC has better structural reconstruction accuracy, compared to the only CTSS algorithm previously available in the literature \citep{nsk03}, when variables in the CTBN can assume more than two values.  For binary variables, that CTSS algorithm performs well, but its performance rapidly degrades as variables are allowed to have increasingly more states. Simulation experiments also showed that the $\chi^2$ test 
for the null state-to-state transition hypothesis has a marginally better performance than the
Kolmogorov-Smirnov test. {However, if we compare the two algorithms in terms of BIC the CTSS
is the best option in all the performed experiments.}
A major limitation of the proposed constraint-based algorithm is the computational cost which becomes problematic in domains with more than 20 variables. However, the the CTSS algorithm has the same limitation.

It would be important to validate the performance of CTPC on real-world data. Unfortunately, we are not aware of any suitable real-world data set where ground truth is available, and thus we were unable to pursue this line of investigation. Furthermore, we are planning additional numerical experiments to evaluate the impact of the type-I-error threshold for the tests to better understand how to calibrate constraint-based algorithms.

\section*{Acknowledgement}
The authors acknowledge the many helpful suggestions of anonymous referees which helped to improve the paper clarity and quality. The authors are indebted to all reviewers for providing extremely useful references to the specialized literature which could impact their future research activity.

\vskip 0.2in
\bibliographystyle{elsarticle-harv} 
\bibliography{pgm2020a}

\begin{appendices}

  \section{CTBN Example: additional cims}
  \label{appendix:extended_cims}
    \begin{table}[H]
      \centering
    \begin{tabular}{|c|c|c|}
    \hline
     & \multicolumn{2}{c|}{Full-Stomach} \\ \hline
    Eating & No & Yes \\ \hline
    \multirow{2}{*}{No} & -0.01 & 0.01 \\ \cline{2-3} 
     & 0.40 & -0.40 \\ \hline
    \end{tabular}
    \quad
    \begin{tabular}{|c|c|c|}
    \hline
     & \multicolumn{2}{c|}{Full-Stomach} \\ \hline
    Eating & No & Yes \\ \hline
    \multirow{2}{*}{Yes} & -2.00 & 2.00 \\ \cline{2-3} 
     & 0.01 & -0.01 \\ \hline
    \end{tabular}
    \caption{$Q_{\text{Full-Stomach}|\text{Eating}}$}
    \end{table}

    \begin{table}[H]
      \centering
    \begin{tabular}{|c|c|c|}
    \hline
     & \multicolumn{2}{c|}{Hungry} \\ \hline
    Full-Stomach & No & Yes \\ \hline
    \multirow{2}{*}{No} & -10.00 & 10.00 \\ \cline{2-3} 
     & 0.01 & -0.01 \\ \hline
    \end{tabular}
    \quad
    \begin{tabular}{|c|c|c|}
    \hline
     & \multicolumn{2}{c|}{Hungry} \\ \hline
    Full-Stomach & No & Yes \\ \hline
    \multirow{2}{*}{Yes} & -0.01 & 0.01 \\ \cline{2-3} 
     & 5.00 & -5.00 \\ \hline
    \end{tabular}
    \caption{$Q_{\text{Hungry}|\text{Full-Stomach}}$}
    \end{table}

  \section{CTPC computational complexity}
  \label{appendix:complexity}

    To compute the time complexity of CTPC, first recall that:
    \begin{itemize}
      \item $\psi$ is the number of transitions occurring in the data set;
      \item $\mathbf{X}$ is the set of nodes of the CTBN;
      \item $n$ is the number of nodes of the CTBN;
      \item $X_k \in \mathbf{X}$ is the node associated with the $k-th$ random variable;
      \item $|X_k|$ is the cardinality of the node $X_k$;
      \item $\gamma = max(\{\mid X_k \mid : X_k \in \mathbf{X}\})$ is the maximum node cardinality present in the network;
      \item $Pa(X_k) \subset \mathbf{X}$ is the parent set of node $X_k$;
      \item $|Pa(X_k)|$ is the size of the paret set $Pa(X_k)$;
      \item $\rho = max(\{\mid Pa(X_k) \mid : X_k \in \mathbf{X}\}) $ it the maximum parent set size present in the network;
      \item $T,\; M$: are the sets of sufficient statistics  $T_{x_i \given \u}$ and $M_{x_i x_m \given \u}$ for a specific node and parent set.
    \end{itemize}

    \begin{figure}[b!]
    \begin{center}
    \includegraphics[width=0.4\linewidth]{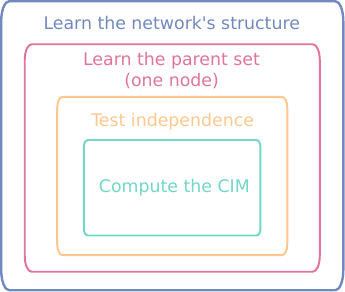}
    \caption{Conceptual model of the CTPC algorithm.}
    \label{complexity_matrioska}
    \end{center}
    \end{figure}

    The structure of the CTPC algorithm can be represented with a series of blocks nested within each other (Figure~\ref{complexity_matrioska}):
    \begin{enumerate}
      \item \textbf{Compute the CIM}: learn the CIM of a node from a data set given a separating set. 
      \item \textbf{Test Independence}: test the independence between two nodes given a separating set.
      \item \textbf{Learn the parent set (one node)}: learn the parent set of a node.
      \item \textbf{Learn the network's structure}: learn the parent set for each node in the network.
    \end{enumerate}
    We derive the complexity of the CTPC algorithm in a bottom-up fashion, that is, by using the complexity of the inner blocks to derive the complexity of the outer blocks.

    \subsection{Compute the CIM}

    This is the innermost box in Figure~\ref{complexity_matrioska}. The tasks requiring the largest number of operations, which determine the leading terms of the computational complexity, are the following:
    \begin{itemize}
      \item Computing the sufficient statistics $T$ and $M$ over the subset of the dataset containing the analyzed node and its parent set: $O((\rho + 1) \cdot \psi )$. 
      \item Computing the CIM $\Q$, given the sufficient statistics $T$ and $M$. For this task we need to compute a matrix of dimension $\gamma ^2$ for each combination of the parent set: $O(\gamma ^ {\rho + 2})$.
    \end{itemize}
    Overall, the time complexity of this box is $O((\rho + 1) \cdot \psi) +\gamma ^ {\rho + 2})$.

    \subsubsection{Test Independence}

    For this box, we characterise separately the best case and the worst case computational complexity. Recall that we perform two tests to assess whether two variables are independent: one for the \textit{null time to transition hypothesis} and one for the \textit{null state-to-state transition hypothesis}. If at least one of these tests fails, we conclude that the two nodes are not independent.
    \begin{itemize}
      \item \textbf{Best Case}: when the first hypothesis test fails. The most complex operations are:
            \begin{itemize}
              \item The computation of the CIM: $O((\rho + 1) \cdot \psi + \gamma ^{\rho + 2})$.
              \item The computation of one test: $O(1)$.
            \end{itemize}
            Since the computational complexity of one test is $O(1)$ we can ignore it in the overall complexity of this case.
      \item \textbf{Worst Case}: when no hypothesis test fails. The most complex operations are:
            \begin{itemize}
              \item The computation of the CIM: $O((\rho + 1) \cdot \psi + \gamma ^{\rho + 2})$.
              \item Performing all the hypothesis tests, $\gamma$ t-tests (each with complexity $O(1)$) and $\gamma$ chi-square tests (each with complexity $O(\gamma)$) for each possible value of the parent set $\gamma ^{\rho}$: $O(\gamma ^{\rho + 1} + \gamma ^{\rho + 2}) = O(\gamma ^ {\rho + 2})$.
            \end{itemize}
            These two operations are performed in sequence. It follows that to identify the overall complexity of this case it is sufficient to add the complexities of the two operations. 
    \end{itemize}
    Therefore, the complexity of the independence test can be summarized as:
    \begin{center}
        \begin{tabular}{c|c}
            \textbf{Case} & \textbf{Complexity}  \\
            \hline
            Best Case  & $O((\rho + 1) \cdot \psi + \gamma ^{\rho + 2})$ \\
            Worst case & $O((\rho + 1) \cdot \psi + 2\cdot \gamma ^{\rho + 2})$ \\
        \end{tabular}
    \end{center}

    \subsection{Learn the parent set (one node)}
    \label{learn_parameter_set_complexity}

    For this box of the algorithm, we distinguish three cases:
    \begin{itemize}
      \item \textbf{Best Case}: the target node has no parents, thus the most complex operation is:
      \begin{itemize}
         \item The independence test (worst case): $O(2 \cdot \gamma ^{\rho + 2} + (\rho + 1) \cdot \psi)$.
         
          If the node under analysis has no parents it follows that $\rho  = 0$  from the definition of $\rho$. 
          We need to test all the possible parent set with size 1 ($\rho + 1$), hence we perform $n-1 \to O(n)$ tests for an overall complexity of $O(n \cdot (2 \cdot\gamma ^3 + 2\cdot \psi))$.
    \end{itemize}

      \item \textbf{Worst case}: all nodes are parents of the current node, thus the most complex operation is:
        \begin{itemize}
          \item The independence test (best case): $O((\rho + 1) \cdot \tau  + \gamma ^{\rho + 2})$. The target node depends on all other nodes ($\rho = n - 1$) and therefore the first independence test always fail. The algorithm computes all potential parent sets, for an overall complexity of 
          \begin{equation*}
            O \left (\sum ^{n - 1} _{r=1} \frac{(n - 1)!}{r! \cdot (n - 1 - r)!} \cdot (\gamma ^ {r + 2} + (r + 1) \cdot \psi) \right)
          \end{equation*}
          which simplifies to $O(2^{n} \cdot (\gamma^{n +1} + n  \cdot \psi))$ since
          \begin{equation*}
           O(\gamma ^ {r + 2} + r   \cdot \psi) \sim O(\gamma^{n} + n \cdot \psi).
          \end{equation*}
          and 
          \begin{equation*}
           \sum ^{n - 1} _{r=1} \frac{(n - 1)!}{r! \cdot (n - 1 - r)!} \to O(2^n)
          \end{equation*}
        \end{itemize}
      \item \textbf{General case}: The General case is similar to the worst case with one important difference: $\rho < n -1$ instead of $\rho = n -1$. The most complex operations are:
        \begin{itemize}
          \item Independence test (best case): $O((\rho + 1) \cdot \psi + \gamma ^ {\rho + 2})$. All the tests until that conditional on the true parent set of size $\rho$ fail. Therefore, the algorithm evaluates all possible parent sets of size $\leq \rho + 1$ , resulting in the following complexity:
          \begin{equation}
            O \left(\sum ^{\rho + 1} _{r=1} \frac{(n - 1)!}{r! \cdot (n - 1 - r)!} \cdot (\gamma ^ {r + 2} + (r + 1) \cdot \psi)\right).
          \label{eq:testgencase}
          \end{equation}
        \end{itemize}
        
        Noting that:
        \begin{align*}
          O(\gamma ^ {r + 2} + (r+1)  \cdot \psi) &\to O(\gamma^{\rho + 3} + (\rho + 2)\cdot \psi), \\ \\
          O\left(\sum ^{\rho + 1} _{r=1} \frac{(n - 1)!}{r! \cdot (n - 1 - r)!}\right) &\to
            O\left(\frac{n^{\rho + 1}}{(\rho + 1)!}\right).&  
      \end{align*}
      the computational complexity in (\ref{eq:testgencase}) simplifies to:
      \begin{equation*}
        O\left(\frac{n^{\rho + 1}}{(\rho + 1)!} \cdot (\gamma ^{\rho + 3} + (\rho + 2)  \cdot \psi )\right)
      \end{equation*}

    \end{itemize}

    \noindent In conclusion, the complexity of the three cases can be summarized as follows:
    \begin{center}
        \begin{tabular}{c|c}
          \textbf{Case} & \textbf{Complexity}  \\
          \hline
          Best Case & $O(n \cdot (2\cdot\gamma ^3 + 2\cdot \psi))$ \\
          Worst case & $O(2^{n} \cdot (\gamma^{n} + n \cdot \psi))$ \\
          General case&   $ O\left(\frac{n^{\rho + 1}}{(\rho + 1)!} \cdot (\gamma ^{\rho + 3} + (\rho + 2)  \cdot \psi )\right)$
        \end{tabular}
    \end{center}

    \subsubsection{Learn the network's structure}
    In Section~\ref{learn_parameter_set_complexity} we presented the computational complexity of
    learning the parent set of a single node. In order to have the complexity of the CTPC algorithm we
    just need to multiply those equations by $n$.  The computational complexity of the CTPC algorithm
    can be summarized as follows:

    \begin{center}
        \begin{tabular}{c|c}
          \textbf{Case} & \textbf{Complexity}  \\
          \hline
          Best Case     & $O(n^2 \cdot (2\cdot\gamma ^3 + 2\cdot \psi))$ \\
          Worst case    & $O(n \cdot 2^{n} \cdot (\gamma^{n} + n \cdot \psi))$ \\
          General case  & $ O\left(\frac{n^{\rho + 2}}{(\rho + 1)!} \cdot (\gamma ^{\rho + 3} + (\rho + 2)  \cdot \psi )\right)$
        \end{tabular}
    \end{center}

    \section{Experiments: comparison between differet hypothesis tests}
    \label{appendix:experiments1}

    We perform a simulation study using a full factorial experimental design over different numbers of nodes $n = \{5, 10, 15, 20\}$,
    network densities $\{0.1, 0.2, 0.3\}$, number of states for the nodes $|\Val(X_i)| = \{2, 3\}$ and different numbers of trajectories $h = \{100, 200, 300\}$. Each trajectory lasts on average for 100 units of 
    time. We generate 10 replicates for each simulation configuration with $n < 10$ and 3 replicates for configurations with $n \geqslant 10$.\footnote{We use the CTBN-RLE package \citep{shelton2010continuous} for CTSS learning. We provide a Python implementation of CTPC, which is available at \url{https://github.com/AlessandroBregoli/ctbn_cba}. CTBN-RLE uses the Bayesian Score with a Gamma prior for the parameters of the exponential distributions and with a Dirichlet prior for the transition probabilities. Hyperparameters are set to their default value, that is, $\tau=1$ for the Gamma prior and $\alpha=1$ for the Dirichlet prior.}

    Results are summarized in Table~\ref{table:score_based} \citep[for the CTSS algorithm in][]{nsk03}, Table~\ref{table:exp_and_chi2} (for CTPC\textsubscript{$\chi^2$}) and Table~\ref{table:exp_and_ks} (for CTPC\textsubscript{KS}). In the case of binary variables, the CTSS algorithm performs better than the proposed constraint-based algorithms for any combination of network density, number of trajectories and number of nodes. CTPC\textsubscript{$\chi^2$} and CTPC\textsubscript{KS} have comparable performance, which is expected since in this case the two algorithms are identical (because the tests are identical, that is, we only test waiting times). However, CTPC\textsubscript{$\chi^2$} and CTPC\textsubscript{KS} have better F1 scores than the CTSS algorithm for ternary variables. 

    CTPC\textsubscript{$\chi^2$} appears to perform marginally better than CTPC\textsubscript{KS} when $n < 20$, but the two algorithms are again comparable when $n = 20$ and $h = 300$. 
    This suggests that CTPC\textsubscript{$\chi^2$} is more sample-efficient than CTPC\textsubscript{KS} with respect to the number of trajectories. 

    However, CTPC\textsubscript{$\chi^2$} and CTPC\textsubscript{KS} scale better than the available CTSS implementation for the algorithm from \citet{nsk03}, which exhausts the 24GiB of memory allocated for the experiment and fails to complete the execution as shown in the last line of Table~\ref{table:score_based}.

    \begin{table}[H]
        \centering\footnotesize
        
        \begin{tabular}{@{}cccc|ccc|cccccc@{}}
        \toprule
        \textbf{Cardinality} & \multicolumn{9}{c}{Binary variables} \\
        \textbf{Network density} & \multicolumn{3}{c}{0.1} & \multicolumn{3}{c}{0.2} & \multicolumn{3}{c}{0.3} \\ 
        \textbf{\# trajectories} & 100 & 200 & 300 & 100 & 200 & 300 & 100 & 200 & 300 \\
        \midrule
        5 & 1.00 & 1.00 & 1.00 & 1.00 & 1.00 & 1.00 & 1.00 & 1.00 & 1.00 \\
        10 & 1.00 & 1.00 & 1.00 & .990 & .990 & .990 & 1.00 & 1.00 & 1.00 \\
        15 & 1.00 & 1.00 & 1.00 & 1.00 & 1.00 & 1.00 & .941 & .993 & 1.00 \\
        20 & .984 & 1.00 & 1.00 & .987 & 1.00 & 1.00 & .850 & .922 & .934 \\
        \midrule
        \textbf{Cardinality} & \multicolumn{9}{c}{Ternary variables} \\
        \midrule
        5 & 1.00 & 1.00 & 1.00 & 1.00 & 1.00 & 1.00 & 1.00 & 1.00 & 1.00 \\
        10 & 1.00 & 1.00 & 1.00 & .949 & .949 & 1.00 & .987 & .934 & .962 \\
        15 & .971 & .983 & 1.00 & .800 & .841 & 1.00 & .541 & .605 & .765 \\
        20 & --- & --- & --- & --- & --- & --- & --- & --- & --- \\ 
        \bottomrule
        \end{tabular}
        \caption{F$_1$-score for the CTSS algorithm.}
        \label{table:score_based}
        
        \vspace{1.5\baselineskip}
        
        \begin{tabular}{@{}cccc|ccc|ccc@{}}
        \toprule
        \textbf{Cardinality} & \multicolumn{9}{c}{Binary variables} \\
        \textbf{Network density} & \multicolumn{3}{c}{0.1} & \multicolumn{3}{c}{0.2} & \multicolumn{3}{c}{0.3} \\
        \textbf{\# trajectories} & 100 & 200 & 300 & 100 & 200 & 300 & 100 & 200 & 300 \\
        \midrule
        5 & .988 & 1.00 & 1.00 &  1.00 & 1.00 & 1.00 & .992 & 1.0 & 1.00 \\
        10 & 1.00 & .988 & 1.00 & .970 & .970 & .970 & .966 & .973 & .967 \\
        15 & .980 & .994 & 1.00 & .949 & .981 & .993 & .830 & .903 & .933 \\
        20 & .968 & .988 & .992 & .935 & .989 & .980 & .787 & .871 & .883 \\
        \midrule
        \textbf{Cardinality} & \multicolumn{9}{c}{Ternary variables} \\
        \midrule
        5 &  .972 & .921 & .909 & .973 & .973 & .973 & .966 & .953 & .979 \\
        10 & .938 & .938 & .950 & .984 & .992 & .992 & .981 & .975 & .970 \\
        15 & .967 & .962 & .967 & .966 & .984 & .984 & .820 & .871 & .887 \\
        20 & .944 & .944 & .939 & .880 & .904 & .913 & .583 & .720 & .761 \\
        \bottomrule
        \end{tabular}
        \caption{F$_1$-score for the CTPC\textsubscript{$\chi^2$} algorithm.}
        \label{table:exp_and_chi2}
        
        \vspace{1.5\baselineskip}
        
        \begin{tabular}{@{}cccc|ccc|ccc@{}}
        \toprule
        \textbf{Cardinality} & \multicolumn{9}{c}{Binary variables} \\
        \textbf{Network density} & \multicolumn{3}{c}{0.1} & \multicolumn{3}{c}{0.2} & \multicolumn{3}{c}{0.3} \\
        \textbf{\# trajectories} & 100 & 200 & 300 & 100 & 200 & 300 & 100 & 200 & 300 \\
        \midrule
        5 & .988 & 1.00 & 1.00 &  1.00 & 1.00 & 1.00 & .992 & 1.0 & 1.00 \\
        10 & 1.00 & .988 & 1.00 & .970 & .970 & .970 & .966 & .973 & .967 \\
        15 & .980 & .994 & 1.00 & .949 & .981 & .993 & .830 & .903 & .933 \\
        20 & .968 & .988 & .992 & .935 & .989 & .980 & .787 & .871 & .883 \\
        \midrule
        \textbf{Cardinality} & \multicolumn{9}{c}{Ternary variables} \\
        \midrule
        5 &  .667 & .667 & .667 & .766 & .771 & .720 & .871 & .802 & .785 \\
        10 & .617 & .623 & .650 & .811 & .775 & .780 & .890 & .886 & .854 \\
        15 & .762 & .782 & .775 & .840 & .863 & .857 & .775 & .855 & .875 \\
        20 & .644 & .624 & .602 & .820 & .852 & .859 & .602 & .704 & .757 \\
        \bottomrule
        \end{tabular}
        \caption{F$_1$-score for the PC\textsubscript{$KS$}CTBN algorithm.}
        \label{table:exp_and_ks}
        \vspace{\baselineskip}

    \end{table}
    
    \section{Experiment: full tables}
    \label{appendix:experiments2}

    \begin{table}[H]
        \centering
        \footnotesize
        \begin{tabular}{ccccc}
        \toprule
        \textbf{Cardinality} & \multicolumn{4}{c}{Binary variables} \\
        \textbf{Network density} &              0.1 &            0.2 &            0.3 &            0.4 \\
        \midrule
        5  &    $1.00 (\pm.000)$ &  $.986 (\pm.045)$ &  $1.00 (\pm.000)$ &  $1.00 (\pm.000)$ \\
        10 &    $1.00 (\pm.000)$ &  $1.00 (\pm.000)$ &  $1.00 (\pm.000)$ &  $1.00 (\pm.000)$ \\
        15 &    $1.00 (\pm.000)$ &  $1.00 (\pm.000)$ &  $1.00 (\pm.000)$ &  $.982 (\pm.030)$ \\
        20 &    $1.00 (\pm.000)$ &  $.996 (\pm.012)$ &  $.972 (\pm.032)$ &  $.860 (\pm.066)$ \\
        \midrule
        \textbf{Cardinality} & \multicolumn{4}{c}{Ternary variables} \\
        \midrule
        5  &     $1.00 (\pm.000)$ &  $1.00 (\pm.000)$ &  $1.00 (\pm.000)$ &  $1.00 (\pm.000)$ \\
        10 &     $1.00 (\pm.000)$ &  $.982 (\pm.056)$ &  $.968 (\pm.078)$ &  $.948 (\pm.061)$ \\
        15 &     $1.00 (\pm.000)$ &  $.964 (\pm.059)$ &  $.805 (\pm.091)$ &  $.526 (\pm.132)$ \\
        20 &     $.995 (\pm.016)$ &  $.903 (\pm.063)$ &  $.562 (\pm.078)$ &  $.108 (\pm.023)$ \\
        \midrule
        \textbf{Cardinality} & \multicolumn{4}{c}{Quaternary variables} \\
        \midrule
        5  &        $1.00 (\pm.000)$ &  $1.00 (\pm.000)$ &  $1.00 (\pm.000)$ &  $1.00 (\pm.000)$ \\
        10 &        $1.00 (\pm.000)$ &  $.971 (\pm.062)$ &  $.887 (\pm.144)$ &  $.765 (\pm.146)$ \\
        15 &        $.987 (\pm.040)$ &  $.856 (\pm.151)$ &  $1.00 (\pm.000)$ &  $.285 (\pm.122)$ \\
        20 &                 - &           - &           - &           - \\
        \bottomrule
        \end{tabular}
        \caption{F$_1$-score for the CTSS algorithm.}
        \label{table:second_score_based}
        
        \vspace{1.5\baselineskip}
        
        \begin{tabular}{ccccc}
        \toprule
        \textbf{Cardinality} & \multicolumn{4}{c}{Binary variables} \\
        \textbf{Network density} &              0.1 &            0.2 &            0.3 &            0.4 \\
        \midrule
        5  &    $1.00 (\pm.000)$ &  $1.00 (\pm.000)$ &  $.992 (\pm.024)$ &  $1.00 (\pm.000)$ \\
        10 &    $1.00 (\pm.000)$ &  $1.00 (\pm.000)$ &  $.990 (\pm.010)$ &  $.984 (\pm.018)$ \\
        15 &    $.998 (\pm.007)$ &  $.985 (\pm.013)$ &  $.957 (\pm.024)$ &  $.892 (\pm.042)$ \\
        20 &    $.996 (\pm.006)$ &  $.967 (\pm.019)$ &  $.888 (\pm.044)$ &  $.698 (\pm.043)$ \\
        \midrule
        \textbf{Cardinality} & \multicolumn{4}{c}{Ternary variables}  \\
        \midrule
        5  &     $.940 (\pm.065)$ &  $.947 (\pm.058)$ &  $.956 (\pm.068)$ &  $.940 (\pm.070)$ \\
        10 &     $.914 (\pm.056)$ &  $.955 (\pm.031)$ &  $.989 (\pm.008)$ &  $.986 (\pm.016)$ \\
        15 &     $.952 (\pm.042)$ &  $.974 (\pm.012)$ &  $.960 (\pm.051)$ &  $.813 (\pm.097)$ \\
        20 &     $.960 (\pm.025)$ &  $.972 (\pm.039)$ &  $.764 (\pm.056)$ &  $.515 (\pm.102)$ \\
        \midrule
        \textbf{Cardinality} & \multicolumn{4}{c}{Quaternary variables} \\
        \midrule
        5  &        $.812 (\pm.096)$ &  $.816 (\pm.070)$ &  $.836 (\pm.058)$ &  $.868 (\pm.065)$ \\
        10 &        $.765 (\pm.096)$ &  $.840 (\pm.042)$ &  $.906 (\pm.046)$ &  $.950 (\pm.038)$ \\
        15 &        $.785 (\pm.064)$ &  $.918 (\pm.029)$ &  $.783 (\pm.047)$ &  $.710 (\pm.085)$ \\
        20 &                 - &           - &           - &           - \\  
        \bottomrule
        \end{tabular}
        \caption{F$_1$-score for the CTPC\textsubscript{$\chi^2$} algorithm.}
        \label{table:second_exp_and_chi2}
    \end{table}

\end{appendices}
\end{document}